\pdfoutput=1

\documentclass[11pt]{article}

\usepackage{acl}

\usepackage{times}
\usepackage{latexsym}
\usepackage{pgfplots}
\usepackage[utf8]{inputenc}
\usepackage{graphicx}
\usepackage[T1]{fontenc}

\usepackage[utf8]{inputenc}

\usepackage{microtype}

\title{MaNLP@SMM4H'22: BERT for Classification of Twitter Posts}

\author{Keshav Kapur \\
  Manipal Institute of Technology \\
  keshav29kapur@gmail.com
  \And Rajitha Harikrishnan \\
  Manipal Institute of Technology \\
  rajithasuja@gmail.com 
  \AND Sanjay Singh \\
  Manipal Institute of Technology \\
  sanjay.singh@manipal.edu}

\begin{document}

\maketitle
\begin{abstract}

The reported work is our straightforward approach for the shared task “Classification of tweets self-reporting age” organized by the  “Social Media Mining for Health Applications (SMM4H)” workshop. This literature describes the approach that was used to build a binary classification system, that classifies the tweets related to birthday posts into two classes namely, exact age(positive class) and non-exact age(negative class). We made two submissions with variations in the preprocessing of text which yielded F1 scores of 0.80 and 0.81 when evaluated by the organizers. 

\end{abstract}

\section{Introduction}

Determining the exact age of an individual is crucial to increasing the use of social media data for research purposes. In this contemporary world, adolescents use social media to the extent that it can have some very severe effects on
their overall well-being if not monitored. Hence, a few applications like Twitter have set up some age restrictions for the well-being of an individual.
They automatically detect the age of an individual
trying to protect them from viewing unnecessary
and harmful content.\newline
In this work, we determine the exact age of an
individual based on their tweets on Twitter. This
helps in validating if a particular user has faked
their age or not. One of the major challenges we
faced while working with the data is that some of
them could tweet about their friend’s or relative’s
birthday which was getting misclassified. We have
used BERT model which helps in the binary classification of our data. While developing our system for this task, we have discovered BERT that outperforms traditional training models.

\section{Methodology}
\subsection{Pre-Processing}
Initially, the organisers provided us with 8,800 training data and 2,200 validation data. This data-set consisted of three fields: tweet id, text of the Tweet Object and annotated binary class label(exact age present/absent). The training and validation data were later combined and it was pre-processed for further development. For pre-processing, we removed URLs, emoticons, hashtags, and mentions using a python package \emph{tweet-preprocessor}. After that we removed: contractions from the tweets, special characters and extra spaces. Then we used python package called \emph{Natural Language Toolkit} for removing the stop words. After these steps, we have further divided the pre-processing into two techniques: pronouns and removing pronouns.
\subsection{Model}
In our model, we have used BERT (\emph{bert-base-uncased}) from the Hugging Face library as a classifier and Softmax as the activation function. In the BERT model, there is an important special token [CLS] which is used as an input for our choice of classifier. We have used the Adam optimizer to fine-tune our BERT model. We trained the model with 4 to 10 epochs which converged after 10 epochs. Learning rate of the optimizer is given by 5e - 5.The batch size used is 32.

\section{Evaluation}
In the validation phase, our model produced satisfactory results with about 90\%. In the test data, 10,000 tweets were provided by the organizers. We have first pre-processed with pronouns and then removed pronouns in the next round of pre-processing. After evaluation, our models generated an F1-score of 0.80 and 0.81. \newline Table 1 shows our evaluation scores for Precision, Recall, and F1-Score as provided by the organizers. Model 1 shows scores of pre-processing with pronouns and Model 2 shows scores of pre-processing with pronouns removed.

\begin{table}
\centering
\begin{tabular}{c|c|c|c}
\hline
\textbf{Model} & \textbf{Precision} & \textbf{Recall} & \textbf{F1-Score}\\
\hline
{Model 1} & {0.839} & {0.780} & {0.808}\\
{Model 2} & {0.771} & {0.870} & {0.818}
\end{tabular}
\caption{Evaluation scores}
\label{tab:accents}
\end{table}
\hfill

\section{Conclusion}

We discussed our approach to fine-tuning our BERT model on Task 4 of the 2022 Social Media Mining for Health applications shared task. As we observe from the results, the given training data was inadequate to train on a BERT model. There was an imbalance in the number of positives and negatives given in our dataset(refer to Figure 1). An interesting observation drawn from this work is that BERT models rely on huge and balanced datasets for learning patterns. Future work might consider collecting more data points for training, fine-tuning our BERT model, and applying other state-of-the-art methods like RoBERTa.

\hfill
 
\begin{tikzpicture}
\begin{axis}[
ybar, ymin=0,
width=6cm, height=6cm, enlarge x limits=0.8,
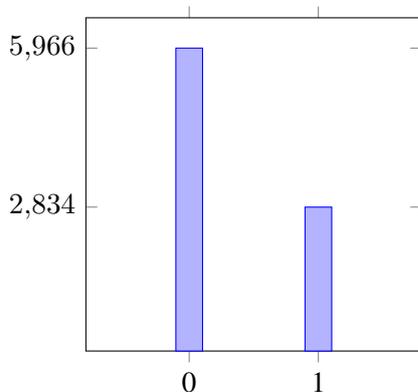
xlabel={Figure 1: Summary of Dataset Labels},
symbolic x coords={0, 1},
ytick=data,
]
\addplot coordinates {(0, 5966) (1, 2834)};
\end{axis}
\end{tikzpicture}

\bibliography{custom}

\begin{thebibliography}{5}
\expandafter\ifx\csname natexlab\endcsname\relax\def\natexlab#1{#1}\fi

\bibitem[{Devlin et~al.(2018)Devlin, Chang, Lee, and
  Toutanova}]{DBLP:journals/corr/abs-1810-04805}
Jacob Devlin, Ming{-}Wei Chang, Kenton Lee, and Kristina Toutanova. 2018.
\newblock \href {http://arxiv.org/abs/1810.04805} {{BERT:} pre-training of deep
  bidirectional transformers for language understanding}.
\newblock \emph{CoRR}, abs/1810.04805.

\bibitem[{Kalyan et~al.(2021)Kalyan, Rajasekharan, and
  Sangeetha}]{https://doi.org/10.48550/arxiv.2108.05542}
Katikapalli~Subramanyam Kalyan, Ajit Rajasekharan, and Sivanesan Sangeetha.
  2021.
\newblock \href {https://doi.org/10.48550/ARXIV.2108.05542} {Ammus : A survey
  of transformer-based pretrained models in natural language processing}.

\bibitem[{Kingma and Ba(2014)}]{https://doi.org/10.48550/arxiv.1412.6980}
Diederik~P. Kingma and Jimmy Ba. 2014.
\newblock \href {https://doi.org/10.48550/ARXIV.1412.6980} {Adam: A method for
  stochastic optimization}.

\bibitem[{Liu et~al.(2019)Liu, Ott, Goyal, Du, Joshi, Chen, Levy, Lewis,
  Zettlemoyer, and Stoyanov}]{https://doi.org/10.48550/arxiv.1907.11692}
Yinhan Liu, Myle Ott, Naman Goyal, Jingfei Du, Mandar Joshi, Danqi Chen, Omer
  Levy, Mike Lewis, Luke Zettlemoyer, and Veselin Stoyanov. 2019.
\newblock \href {https://doi.org/10.48550/ARXIV.1907.11692} {Roberta: A
  robustly optimized bert pretraining approach}.

\bibitem[{Magge et~al.(2021)Magge, Klein, Miranda-Escalada, Al-garadi, Alimova,
  Miftahutdinov, Farre-Maduell, Lopez, Flores, O'Connor, Weissenbacher,
  Tutubalina, Sarker, Banda, Krallinger, and
  Gonzalez-Hernandez}]{smm4h-2021-social}
Arjun Magge, Ari Klein, Antonio Miranda-Escalada, Mohammed~Ali Al-garadi,
  Ilseyar Alimova, Zulfat Miftahutdinov, Eulalia Farre-Maduell, Salvador~Lima
  Lopez, Ivan Flores, Karen O'Connor, Davy Weissenbacher, Elena Tutubalina,
  Abeed Sarker, Juan~M Banda, Martin Krallinger, and Graciela
  Gonzalez-Hernandez, editors. 2021.
\newblock \href {https://aclanthology.org/2021.smm4h-1.0} {\emph{Proceedings of
  the Sixth Social Media Mining for Health ({\#}SMM4H) Workshop and Shared
  Task}}. Association for Computational Linguistics, Mexico City, Mexico.

\end{thebibliography}
\nocite{DBLP:journals/corr/abs-1810-04805, https://doi.org/10.48550/arxiv.1412.6980, https://doi.org/10.48550/arxiv.1907.11692, smm4h-2021-social, https://doi.org/10.48550/arxiv.2108.05542}

\end{document}